\documentclass[11pt,a4paper]{article}

\usepackage[]{acl}
\usepackage{times}
\usepackage{latexsym}
\usepackage{bbm}
\usepackage[T1]{fontenc}
\usepackage{CJKutf8}

\usepackage{microtype}

\usepackage{times}
\usepackage{latexsym}
\usepackage{color}
\usepackage{pifont}
\usepackage{makecell}

\usepackage{stackengine}

\usepackage{latexsym}
\usepackage{amsthm,amsmath,amssymb}
\usepackage{multirow}
\usepackage{array}
\usepackage{mathrsfs}
\usepackage{graphicx}
\usepackage{floatrow}
\usepackage{booktabs}
\usepackage{algorithm}
\usepackage[normalem]{ulem}
\useunder{\uline}{\ul}{}
\usepackage{tikz}
\usepackage{pgfplots}
\usepackage{subcaption}

\usepackage{verbatim}
\usepackage{amsfonts,amssymb}
\usepackage{graphicx} 

\title{SDCL: Self-Distillation Contrastive Learning for Chinese Spell Checking}

\author{Xiaotian Zhang, Hang Yan, Yu Sun, Xipeng Qiu\thanks{*Corresponding author.}\\
    Shanghai Key Laboratory of Intelligent Information Processing, Fudan University \\ 
        School of Computer Science, Fudan University \\
        \{xiaotianzhang21, yusun21\}@m.fudan.edu.cn, \{hyan19, xpqiu\}@fudan.edu.cn
}

\begin{document}
\maketitle
\begin{abstract}
  Due to the ambiguity of homophones, Chinese Spell Checking (CSC) has widespread applications. 
  Existing systems typically utilize BERT for text encoding. 
  However, CSC requires the model to account for both phonetic and graphemic information. 
  %Hence, it does not correspond to the pre-training objective of BERT. 
  To adapt BERT to the CSC task, we propose a token-level self-distillation contrastive learning method.
  We employ BERT to encode both the corrupted and corresponding correct sentence. 
  %Then, contrastive loss pulls the distance between the representations of the perturbed token and the correct token, enabling the model to learn phonetic or graphic information. 
  Then, we use contrastive learning loss to regularize corrupted tokens' hidden states to be closer to counterparts in the correct sentence.
  On three CSC datasets, we confirmed our method provides a significant improvement above baselines.
\end{abstract}

% (1) We found that BERT with mlm head is a very strong basline for CSC.
% (2) We propose a simple but effective self-teaching approach to further enhance the performance. 

\section{Introduction}
Chinese Spell Checking (CSC) is a task to detect and correct spelling mistakes in Chinese sentences. It differs from Chinese Grammatical Error Diagnosis (CGED) in that CSC will not delete or insert any characters. Because CSC is usually the pre-process of downstream Natural Language Processing (NLP) tasks, CSC has been extensively studied recently \cite{hong2019faspell,DBLP:conf/acl/ChengXCJWWCQ20,DBLP:conf/emnlp/JiYQ21,DBLP:conf/acl/XuLZLWCHM21}. 

\begin{figure}[t]
    \includegraphics[width=1\textwidth]{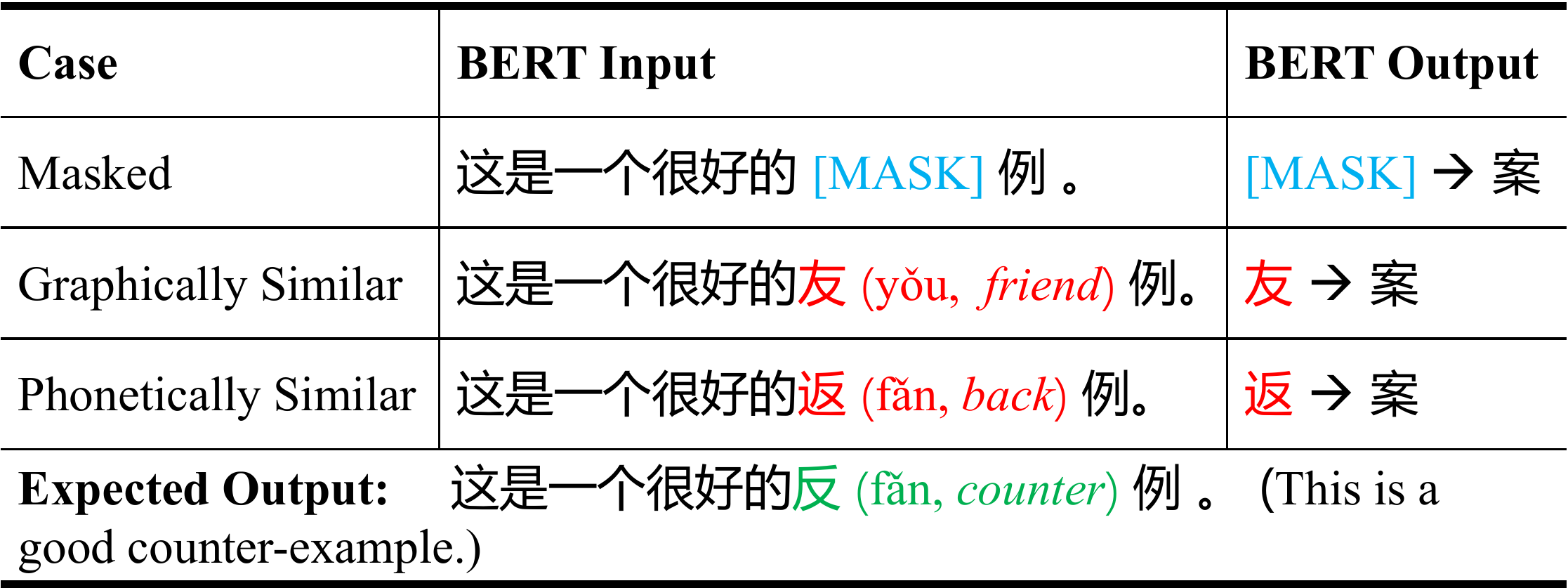}
    \caption{Example of the CSC task. BERT cannot take advantage of the erroneous characters ``友'' (visually similar to ``反'') or ``返''(phonetically similar to ``反'') to recover the expected character ``反''. }\label{fig:example}
  \end{figure}

Since the introduction of pre-trained model BERT \cite{DBLP:conf/naacl/DevlinCLT19}, many works have tried to utilize the power of pre-training models to achieve good results \cite{hong2019faspell,zhang2020spelling}. BERT used the masked language modeling (MLM) task to pre-train, which forces BERT to utilize the contextual information to recover the masked token. The pre-training task makes BERT suitable to conduct predictions through semantic information. However, as pointed out in previous work \cite{DBLP:conf/coling/LiuLCL10}, 83\% and 48\% of the CSC errors are related to phonological similarity and visual similarity, respectively. Therefore, directly using BERT to tackle this task will cause a mismatch between what BERT excels at and what this task needs. An example is depicted in Figure \ref{fig:example}. 
% 中文纠错任务在过去几年得到了大量的研究。最近的工作几乎都依赖了pretrain模型来进行CSC任务\cite{}。但是最近越来越多的论文显示，BERT的表示只是集中在了比较小的圆锥之中。

\begin{CJK}{UTF8}{gbsn}
  \end{CJK}

Previous works have tried to narrow this gap by combining confusion set\footnote{Confusion set contains characters sound or looks like each other.} into their model \cite{DBLP:conf/acl/ChengXCJWWCQ20,hong2019faspell}.  \citet{DBLP:conf/emnlp/JiYQ21,zhang2021correcting,li2021exploration,liu2021plome,DBLP:conf/acl/XuLZLWCHM21} took one step further through adding phonetic or graphic information during the pre-training phase. 

%However, the explicit introduction of the prior information methods only effect the token embedding by cross-entropy loss while adding additional training expenses, as shown in Figure 2. 
However, these methods of explicitly introducing prior information rely on additional training data or parameters, increasing the cost of training. And cross-entropy loss only changes the token embedding of the word, not the contextual embedding, as shown in Figure 2.

Instead of relying on confusion sets or further pre-training, we utilize contrastive learning (CL) to narrow the gap, by pulling together the hidden states of wrong and right character usages. 

Contrastive learning gains great popularity recently for its outstanding performance in learning better image representations\cite{DBLP:conf/icml/ChenK0H20,DBLP:conf/cvpr/He0WXG20}. In NLP, \citet{DBLP:conf/emnlp/GaoYC21,DBLP:conf/iclr/GunelDCS21} tried to use CL to get better sentence representation, the formerly used dropout to form positive samples, and the latter used samples with the same class as positive samples. However, these attempts are mainly focused on the sentence-level, \citet{DBLP:journals/corr/abs-2111-04198} proposed to use the method to adapt the CL in the token-level, and it forms positive token samples by not masking out this token. To reduce training costs, We use an efficient self-distillation method to obtain the positive and negative samples\cite{gao2021disco}\footnote{Since the input and output formulation of the CSC task and the pre-training MLM task is very similar, we can directly use out-of-the-box BERT without adding or deleting any parameters. We input the corrupted sentences into BERT and get predictions for each token, and this process is very similar to the MLM pre-training task, only differs in that we do not mask any tokens. If the output token prediction is different from its input token, we deem BERT detect errors and correct this error as its output token.}.

In summary, %As discussed before, the innate ability of BERT does not align well with the CSC task. \footnote{Since the input and output formulation of the CSC task and the pre-training MLM task is very similar, we can directly use out-of-the-box BERT without adding or deleting any parameters. We input the corrupted sentences into BERT and get predictions for each token, and this process is very similar to the MLM pre-training task, only differs in that we do not mask any tokens. If the output token prediction is different from its input token, we deem BERT detect errors and correct this error as its output token.} 
%Previous work emphasized the modeling phonetical or visual similarity among characters~\cite{DBLP:conf/acl/XuLZLWCHM21}. 
%However, finding the clusters to which characters belong doesn't make them any easier to distinguish. 
%However, the explicit introduction of the prior information can not directly help model distinguish errors.
%The phenomenon is depicted in Figure \ref{fig:respresentation}. 
%And too tight correlations may harm the model performance. 
We propose a Self-Distillation Contrastive Learning (SDCL) method to alleviate the phenomenon.
%Specifically, we use regularize to help BERT learn uniformly contextual embedding distribution for CSC task. 
Without extra parameters and training data, our method use regularization loss to help BERT learn uniformly contextual embedding distribution, achieve significant performance gains on the baseline, and can even surpass specific pre-trained models achieving comparable performance to SOTA in three CSC datasets. 

% CL最近受到了很大的关注，因为它能够在无监督的

% \cite{} xxxx. While to further enhance the performace \cite{} tried to incorporate Pinyin and strokes during the pre-training process. 

% Instead of relying on confusion sets or further pre-training, we propose to use contrastive loss to narrow this gap. 

% Although adding Pinyin and strokes into the finetuning or pre-training processing of BERT is quite straightforward, it makes the model more complex and adding more computational overheads. 

\begin{figure}[t]
  \includegraphics[width=1\textwidth]{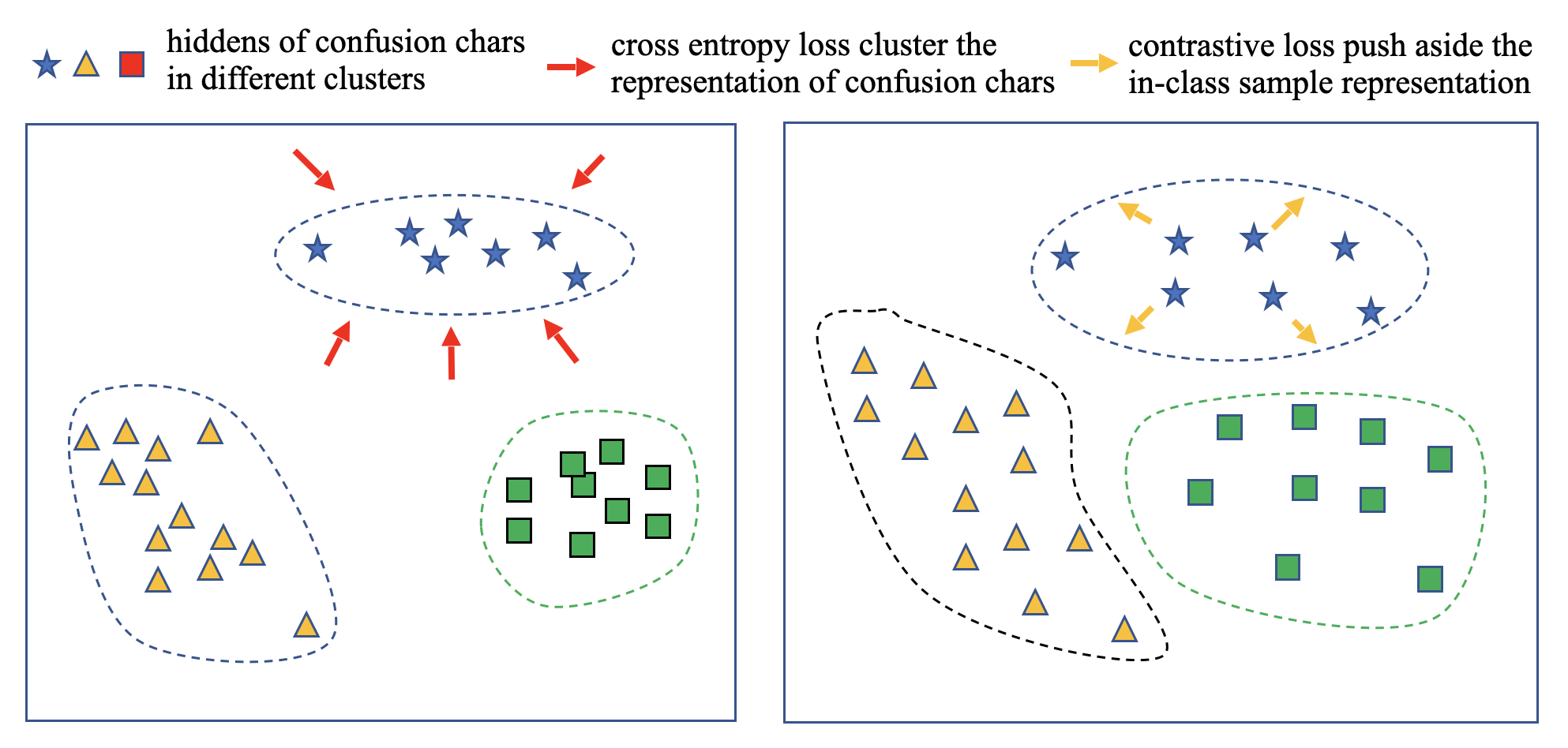}
  \caption{The visualization of confusion samples with the same context, using bert as an encoder. The feature
modeling by cross-entropy loss only clusters the correction pairs. While contrastive loss makes model tradeoffs alignment and uniformity.
  }\label{fig:respresentation}
\end{figure}

\begin{table*}[!ht]
  \centering
  \resizebox{\textwidth}{72mm}{
  \begin{tabular}{clcccccc}
  \hline
  \multicolumn{1}{|c|}{\multirow{2}{*}{\textbf{Dataset}}} &
    \multicolumn{1}{c|}{\multirow{2}{*}{\textbf{Method}}} &
    \multicolumn{3}{c|}{\textbf{Detection Level}} &
    \multicolumn{3}{c|}{\textbf{Correction Level}} \\ \cline{3-8} 
  \multicolumn{1}{|c|}{} &
    \multicolumn{1}{c|}{} &
    \multicolumn{1}{c|}{Pre} &
    \multicolumn{1}{c|}{Rec} &
    \multicolumn{1}{c|}{F1} &
    \multicolumn{1}{c|}{Pre} &
    \multicolumn{1}{c|}{Rec} &
    \multicolumn{1}{c|}{F1} \\ \hline
    \multicolumn{1}{|c|}{\multirow{10}{*}{SIGHAN13}} &
    \multicolumn{1}{l|}{FASpell\cite{hong2019faspell}} &
    76.2 &
    63.2 &
    \multicolumn{1}{c|}{69.1} &
    73.1 &
    60.5 &
    \multicolumn{1}{c|}{66.2} \\
  \multicolumn{1}{|c|}{} &
    \multicolumn{1}{l|}{SpellGCN\cite{DBLP:conf/acl/ChengXCJWWCQ20}} &
    80.1 &
    74.4 &
    \multicolumn{1}{c|}{77.2} &
    78.3 &
    72.7 &
    \multicolumn{1}{c|}{75.4} \\
  \multicolumn{1}{|c|}{} &
    \multicolumn{1}{l|}{BERT\cite{DBLP:conf/acl/XuLZLWCHM21}} &
    79.0 &
    72.8 &
    \multicolumn{1}{c|}{75.8} &
    77.7 &
    71.6 &
    \multicolumn{1}{c|}{74.6} \\ \cline{2-8}
  \multicolumn{1}{|c|}{} &
    \multicolumn{1}{l|}{$\text{MLM-phonetics}^{\bigstar}$
    \cite{zhang2021correcting}} &
    82.0 &
    78.3 &
    \multicolumn{1}{c|}{80.1} &
    79.5 &
    77.0 &
    \multicolumn{1}{c|}{78.2}  \\
  \multicolumn{1}{|c|}{} &
    \multicolumn{1}{l|}{$\text{BERT +  Adversarial training}^{\spadesuit}$ \cite{li2021exploration}} &
    - &
    - &
    \multicolumn{1}{c|}{84.0} &
    - &
    - &
    \multicolumn{1}{c|}{83.5} \\
  \multicolumn{1}{|c|}{} &
    \multicolumn{1}{l|}{$\text{REALISE}^{\clubsuit}$
    \cite{DBLP:conf/acl/XuLZLWCHM21}} &
    88.6 &
    82.5 &
    \multicolumn{1}{c|}{85.4} &
    87.2 &
    81.2 &
    \multicolumn{1}{c|}{84.1} \\ 
  \multicolumn{1}{|c|}{} &
    \multicolumn{1}{l|}{$\text{BERT + Pre-trained for CSC}^{\spadesuit}$
    \cite{li2021exploration}} &
    - &
    - &
    \multicolumn{1}{c|}{84.9} &
    - &
    - &
    \multicolumn{1}{c|}{84.4} \\ \cline{2-8} 
  \multicolumn{1}{|c|}{} &
    \multicolumn{1}{l|}{SDCL(ours)} &
    \multicolumn{1}{l}{$\text{\textbf{88.9}}^{\uparrow}$} &
    \multicolumn{1}{l}{$\text{\textbf{81.8}}^{\uparrow}$} &
    \multicolumn{1}{l|}{$\text{\textbf{85.2}}^{\uparrow}$} &
    \multicolumn{1}{l}{$\text{\textbf{88.0}}^{\uparrow}$} &
    \multicolumn{1}{l}{$\text{\textbf{81.0}}^{\uparrow}$} &
    \multicolumn{1}{l|}{$\text{\textbf{84.3}}^{\uparrow}$} \\
     \multicolumn{1}{|c|}{} &
    \multicolumn{1}{l|}{\quad w/o CL(ablation)} &
    80.2 &
    75.7 &
    \multicolumn{1}{l|}{77.9} &
    80.4 &
    73.7 &
    \multicolumn{1}{l|}{76.9} \\ \hline
  \multicolumn{1}{|c|}{\multirow{10}{*}{SIGHAN14}} &
    \multicolumn{1}{l|}{Hybrid\cite{wang2018hybrid}} &
    51.9 &
    66.2 &
    \multicolumn{1}{c|}{58.2} &
    - &
    - &
    \multicolumn{1}{c|}{56.1} \\
  \multicolumn{1}{|c|}{} &
    \multicolumn{1}{l|}{FASpell\cite{hong2019faspell}} &
    61.0 &
    53.5 &
    \multicolumn{1}{c|}{57.0} &
    59.4 &
    - &
    \multicolumn{1}{c|}{-} \\
  \multicolumn{1}{|c|}{} &
    \multicolumn{1}{l|}{SpellGCN\cite{DBLP:conf/acl/ChengXCJWWCQ20}} &
    65.1 &
    69.5 &
    \multicolumn{1}{c|}{67.2} &
    63.1 &
    67.2 &
    \multicolumn{1}{c|}{65.3} \\
  \multicolumn{1}{|c|}{} &
    \multicolumn{1}{l|}{BERT\cite{DBLP:conf/acl/XuLZLWCHM21}} &
    64.5 &
    68.6 &
    \multicolumn{1}{c|}{66.5} &
    62.4 &
    66.3 &
    \multicolumn{1}{c|}{64.3} \\ \cline{2-8} 
  \multicolumn{1}{|c|}{} &
    \multicolumn{1}{l|}{$\text{BERT +  Adversarial training}^{\spadesuit}$\cite{li2021exploration}} &
    - &
    - &
    \multicolumn{1}{c|}{68.4} &
    - &
    - &
    \multicolumn{1}{c|}{66.8} \\
  \multicolumn{1}{|c|}{} &
    \multicolumn{1}{l|}{$\text{REALISE}^{\clubsuit}$ \cite{DBLP:conf/acl/XuLZLWCHM21}} &
    67.8 &
    71.5 &
    \multicolumn{1}{c|}{69.6} &
    66.3 &
    70.0 &
    \multicolumn{1}{c|}{68.1} \\
  \multicolumn{1}{|c|}{} &
    \multicolumn{1}{l|}{$\text{BERT + Pre-trained for CSC}^{\spadesuit}$\cite{li2021exploration}} &
    - &
    - &
    \multicolumn{1}{c|}{70.4} &
    - &
    - &
    \multicolumn{1}{c|}{68.6} \\
  \multicolumn{1}{|c|}{} &
    \multicolumn{1}{l|}{$\text{MLM-phonetics}^{\bigstar}$\cite{zhang2021correcting}} &
    66.2 &
    73.8 &
    \multicolumn{1}{c|}{69.8} &
    64.2 &
    73.8 &
    \multicolumn{1}{c|}{68.7} \\ \cline{2-8} 
  \multicolumn{1}{|c|}{} &
    \multicolumn{1}{l|}{SDCL(ours)} &
    \multicolumn{1}{l}{$\text{\textbf{69.7}}^{\uparrow}$} &
    \multicolumn{1}{l}{\textbf{70.3}} &
    \multicolumn{1}{l|}{$\text{\textbf{70.0}}$} &
    \multicolumn{1}{l}{$\text{\textbf{70.2}}^{\uparrow}$} &
    \multicolumn{1}{l}{\textbf{67.5}} &
    \multicolumn{1}{l|}{$\text{\textbf{68.8}}^{\uparrow}$} \\
      \multicolumn{1}{|c|}{} &
    \multicolumn{1}{l|}{\quad w/o CL(ablation) }  &
    65.8 &
    69.3 &
    \multicolumn{1}{l|}{67.5} &
    64.1 &
    67.6 &
    \multicolumn{1}{l|}{65.8} \\ \hline
  \multicolumn{1}{|r|}{\multirow{12}{*}{SIGHAN15}} &
    \multicolumn{1}{l|}{Hybrid\cite{wang2018hybrid}} &
    56.6 &
    69.4 &
    \multicolumn{1}{c|}{62.3} &
    - &
    - &
    \multicolumn{1}{c|}{57.1} \\
  \multicolumn{1}{|r|}{} &
    \multicolumn{1}{l|}{FASpell\cite{hong2019faspell}} &
    67.6 &
    60.0 &
    \multicolumn{1}{c|}{63.5} &
    66.6 &
    59.1 &
    \multicolumn{1}{c|}{62.6} \\
  \multicolumn{1}{|r|}{} &
    \multicolumn{1}{l|}{Soft-Masked BERT\cite{zhang2020spelling}} &
    \multicolumn{1}{l}{73.7} &
    \multicolumn{1}{l}{73.2} &
    \multicolumn{1}{l|}{73.5} &
    \multicolumn{1}{l}{66.7} &
    \multicolumn{1}{l}{66.2} &
    \multicolumn{1}{l|}{66.4} \\
  \multicolumn{1}{|r|}{} &
    \multicolumn{1}{l|}{SpellGCN\cite{DBLP:conf/acl/ChengXCJWWCQ20}} &
    74.8 &
    \textbf{80.7} &
    \multicolumn{1}{c|}{77.7} &
    72.1 &
    \textbf{77.7} &
    \multicolumn{1}{c|}{75.9} \\ 
  \multicolumn{1}{|r|}{} &
    \multicolumn{1}{l|}{BERT\cite{DBLP:conf/acl/XuLZLWCHM21}} &
    74.2 &
    78.0 &
    \multicolumn{1}{c|}{76.1} &
    71.6 &
    75.3 &
    \multicolumn{1}{c|}{73.4} \\ \cline{2-8} 
  \multicolumn{1}{|r|}{} &
    \multicolumn{1}{l|}{$\text{PLOME}^{\spadesuit}$\cite{liu2021plome}} &
    77.4 &
    81.5 &
    \multicolumn{1}{c|}{79.4} &
    75.3 &
    79.3 &
    \multicolumn{1}{c|}{77.2} \\
  \multicolumn{1}{|r|}{} &
    \multicolumn{1}{l|}{$\text{MLM-phonetics}^{\bigstar}$\cite{zhang2021correcting}} &
    77.5 &
    83.1 &
    \multicolumn{1}{c|}{80.2} &
    74.9 &
    80.2 &
    \multicolumn{1}{c|}{77.5} \\
  \multicolumn{1}{|r|}{} &
    \multicolumn{1}{l|}{$\text{REALISE}^{\clubsuit}$\cite{DBLP:conf/acl/XuLZLWCHM21}} &
    77.3 &
    81.3 &
    \multicolumn{1}{c|}{79.3} &
    75.9 &
    79.9 &
    \multicolumn{1}{c|}{77.8} \\
  \multicolumn{1}{|r|}{} &
    \multicolumn{1}{l|}{$\text{BERT + Pre-trained for CSC}^{\spadesuit}$\cite{li2021exploration}} &
    - &
    - &
    \multicolumn{1}{c|}{79.8} &
    - &
    - &
    \multicolumn{1}{c|}{78.0} \\
  \multicolumn{1}{|r|}{} &
    \multicolumn{1}{l|}{$\text{BERT +  Adversarial training}^{\spadesuit}$ \cite{li2021exploration}} &
    - &
    - &
    \multicolumn{1}{c|}{80.0} &
    - &
    - &
    \multicolumn{1}{c|}{78.2} \\ \cline{2-8} 
\multicolumn{1}{|c|}{} &
    \multicolumn{1}{l|}{SDCL(ours)} &
    $\text{\textbf{81.2}}^{\uparrow}$ &
    79.1 &
    \multicolumn{1}{c|}{$\text{\textbf{80.1}}$} &
    $\text{\textbf{79.3}}^{\uparrow}$ &
    77.5 &
    \multicolumn{1}{c|}{$\text{\textbf{78.3}}^{\uparrow}$} \\
    \multicolumn{1}{|c|}{} &
    \multicolumn{1}{l|}{\quad w/o CL(ablation)} &
    75.9 &
    80.0 &
    \multicolumn{1}{c|}{77.9} &
    74.1 &
    78.0 &
    \multicolumn{1}{c|}{ 76.0 } \\ \hline
  \multicolumn{1}{l}{} &
     &
    \multicolumn{1}{l}{} &
    \multicolumn{1}{l}{} &
    \multicolumn{1}{l}{} &
    \multicolumn{1}{l}{} &
    \multicolumn{1}{l}{} &
    \multicolumn{1}{l}{}
  \caption{Main results of our model. The " $^{\spadesuit}$ " symbol means additional training data, "$^{\clubsuit}$" symbol indicates extra model parameters, and "$^{\bigstar}$" symbol means both. ${\uparrow}$ hints our method performs a significant test $p \mbox{-} value < 0.05$ when comparing with baseline.
  } \label{tb:main}
  
  \end{tabular}}
\end{table*}

\section{Methodology}

We first introduce the formulation of the CSC task, then present the details of our proposed Self-Distillation Contrastive Learning model.
\subsection{The Main Model}
The CSC task can be formulated as given an input sentence $X=[x_1, ..., x_n]$ with $n$ characters, the model needs to output its corresponding correct sentence $Y=[y_1, ..., y_n]$. Usually, most tokens in $Y$ will be the same as their counterpart in $X$. 

\begin{figure}[t]
  \includegraphics[width=1\textwidth]{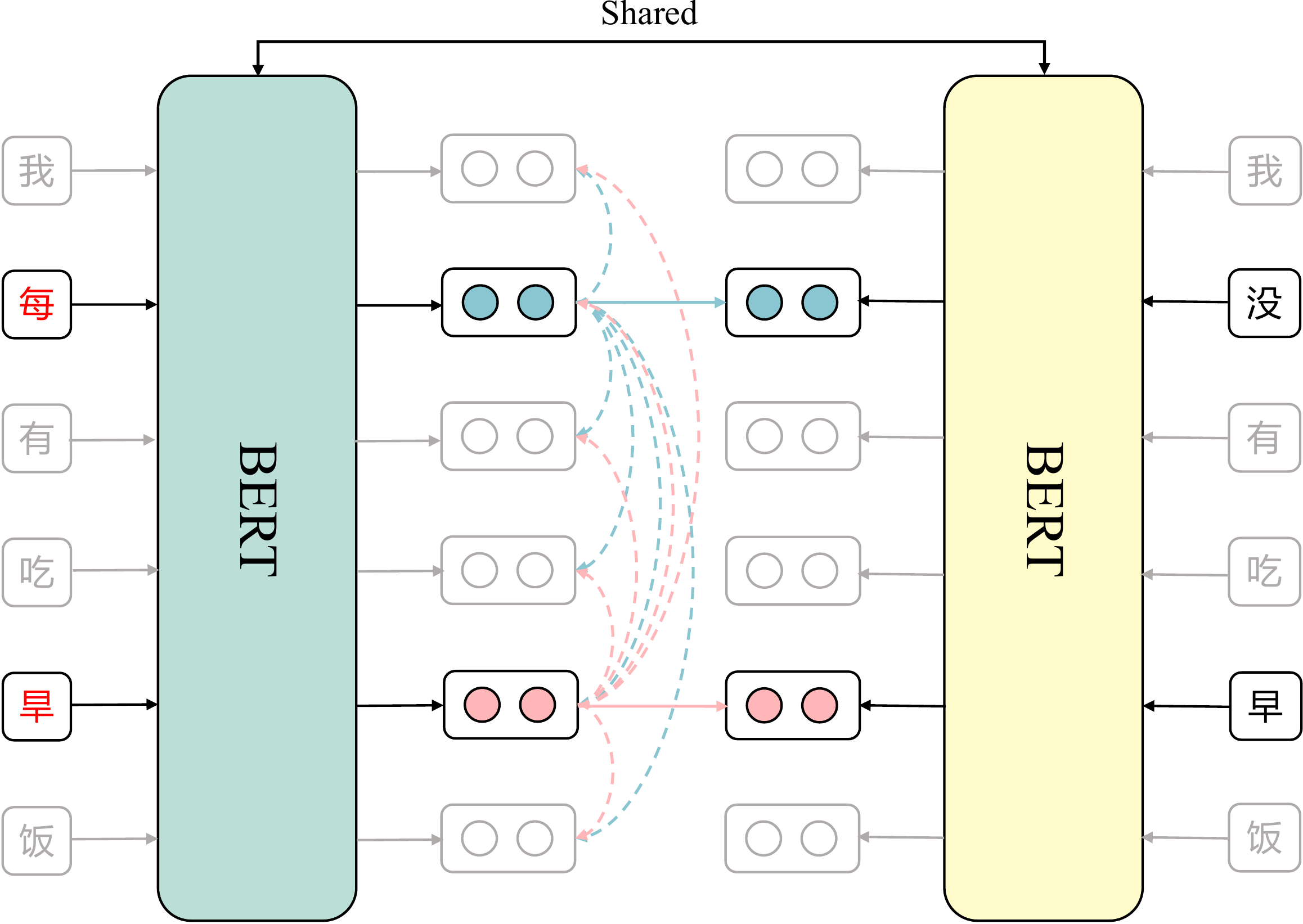}
  \caption{The contrastive loss for our model. We send the wrong and right sentences into the left and right BERT, respectively. We deem each wrong character and its corresponding right character from the right teacher BERT as the positive sample (dense lines), and other characters from the left BERT as negative samples (dotted lines). Lines of different colors correspond to errors at different positions.
  }\label{fig:model}
\end{figure}

We use $\mathrm{MacBERT}$ \cite{cui2020revisiting} as a strong backbone to extract the semantic features of $X$ and then use dot products with the word embedding $\mathbf{W}$ to output the character distribution. This process can be formulated as 
%masked language model head of the pre-trained BERT 
\begin{align}
  \mathbf{H} & = \mathrm{BERT}(X), \nonumber\\
  P(\hat{Y}|X) & = \mathrm{softmax}(\mathbf{W \cdot H}), \nonumber
  %P(\hat{Y}|X) & = \mathrm{MLMHead}(\mathbf{H}), \nonumber
\end{align}
where $\mathrm{BERT()}$ takes the sentence $X$ as input, and outputs a contexualized dense matrix $\mathbf{H} \in \mathbb{R}^{n \times d}$ and $d$ is the hidden state dimension;
% $\mathrm{MLMHead()}$ takes in a vector $\mathbf{H}_i$, and output the token distribution $P(\hat{Y}|X) \in \mathbb{R} ^ {n \times |V|}$, $|V|$ is the vocabulary size. 
We use dot product to calculate the similarity of $\mathbf{W_i}$ and $\mathbf{H}_i$ rather than $\mathrm{MLMHead()}$ (Since there is little performance difference and we set this for implement contrastice loss), then use the result after $\mathrm{softmax}$ as the token distribution $P(\hat{Y}|X) \in \mathbb{R} ^ {n \times |V|}$, $|V|$ is the vocabulary size. 
After getting the token distribution, we calculate the cross-entropy loss as follows

\begin{align}
  \mathcal{L}_x = - \sum_i^n \mathrm{log}(P(\hat{Y}_i=y_i|X)) \nonumber
\end{align}

\subsection{Contrastive Loss}
As mentioned before, most of the errors in the CSC task are caused by phonological or visual similarities, instead of semantical misuses. Therefore, it is hard for BERT to recover the right characters based on contextualized information. 

We propose using an extra loss to help BERT build the connection between the erroneous characters and their corresponding right ones. We want the BERT model to output the close hidden states for the corrupted sentence and its corresponding right sentence through this loss. We propose a self-distillation method with a shared weights teacher BERT to construct positive samples for contrastive learning.
% We fix the parameters of the $\mathrm{MLMHead}$.  
Specifically, the calculation of the loss is as follows

\begin{align}
  -\sum_{i=1}^{n} \mathbbm{1}\left(\tilde{x}_{i}\right) \log \frac{\exp \left(\operatorname{sim}\left(\tilde{h}_{i}, h_{i}\right) / \tau \right)}{\sum_{j=1}^{n} \exp \left(\operatorname{sim}\left(\tilde{h}_{i}, h_{j}\right) / \tau \right)}
  \nonumber
\end{align}

where $\mathbbm{1}\left(\tilde{x}_{i}\right) = 1$ if $x_i$ is the error token, else $0$. The $\tilde{h}_{i}$ indicates the corresponding hiddens from teacher BERT with golden input; $\tau$ is the temperature hyper-parameter and $\mathrm{Sim}(\tilde{h}_{i}, h_{i})$ is the cosine similarity between these two vectors. Minimizing $\mathcal{L}_c$ aims to make the hidden states of the corrupted tokens similar to their correct counterpart. We sample from the batch as negative samples rather than confusion set~\cite{DBLP:conf/acl-sighan/WuLL13} to improve training speed.
We also add a cross-entropy loss for the teacher BERT to repeat the inputs. We use stop gradient (sg) to decouple the gradient backpropagation to $\tilde{h}_{i}$ for stability during training and the final loss is as follows 
%\footnote{Since in the pre-training process of BERT, some of the tokens will be randomly replaced by other tokens. Therefore, even when provided with correct sentences, BERT still predicts tokens not appear in the input.}

\begin{align}
  \mathcal{L}_y & = - \sum_i^n \mathrm{log}(P(\bar{Y}_i=y_i|Y')), \nonumber \\
  \mathcal{L} & = \mathcal{L}_x + \alpha \mathcal{L}_y + \beta \mathcal{L}_c \nonumber, 
\end{align}
where $\alpha, \beta$ is the hyper-parameter. The general model structure is depicted in Figure \ref{fig:model}.

\section{Experiments}
% We will present the experimental setups and main results in this section. 

\subsection{Data and Metrics}
To show the effectiveness of our proposed method, we conduct experiments in three CSC datasets, namely SIGHAN13\cite{DBLP:conf/acl-sighan/WuLL13} SIGHAN14\cite{DBLP:conf/acl-sighan/YuLTC14}, SIGHAN15~\cite{DBLP:conf/acl-sighan/TsengLCC15}. We use the sentence-level metric for both detection and correction to evaluate~\cite{DBLP:conf/acl/ChengXCJWWCQ20}. Our settings are consistent with previous work~\cite{DBLP:conf/acl/XuLZLWCHM21}. More details on data, metrics, and implementation can be found in the Appendix.

\subsection{Main Results}
The main experimental results are depicted in Table \ref{tb:main}. %The ``backbone'' has the same model architecture as ``SDCL'', but its $\alpha$ is set to 0, 
%which means the only difference between ``backbone'' and ``SDCL'' is whether to use the contrastive loss. 
Results show that adding contrastive loss consistently enhances the performance in three datasets. Moreover, our ``SDCL'' even surpasses various further pre-trained models.

% TODO 增加一个我们的baseline是没有contrastive loss的部分

% 首先介绍一下实验设定
% 然后呈现一下主要的results
% 现在面临的困境：（1）SIGHAN等的数据不一致的问题；（2）OCR等用的是官方评测工具来进行评测。

% OCR来自于faspell

\section{Analysis}
In order to show that our model can help BERT correct phonetically or visually similar errors, we design two probing tasks. 

\begin{figure}
        \begin{subfigure}[b]{0.5\textwidth}
                \centering
                \includegraphics[width=.85\linewidth]{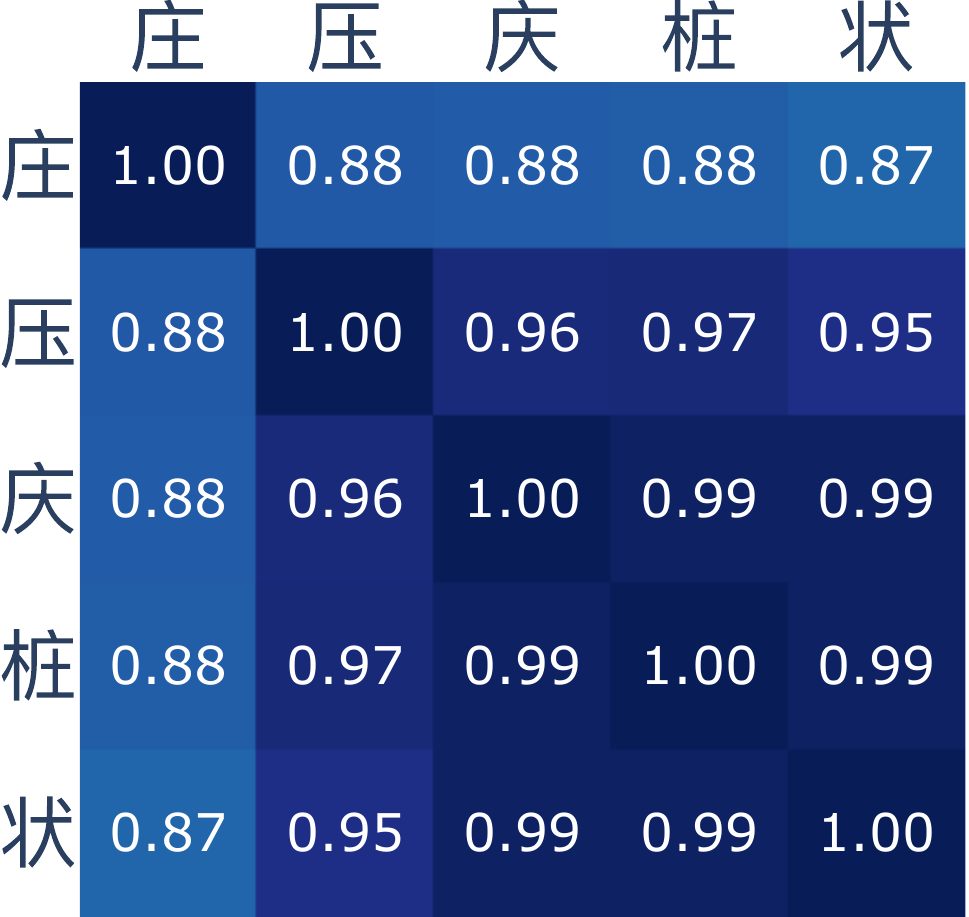}
                \caption{Vanilla BERT}
                \label{fig:gull}
        \end{subfigure}%
        \begin{subfigure}[b]{0.6\textwidth}
                \centering
                \includegraphics[width=.85\linewidth]{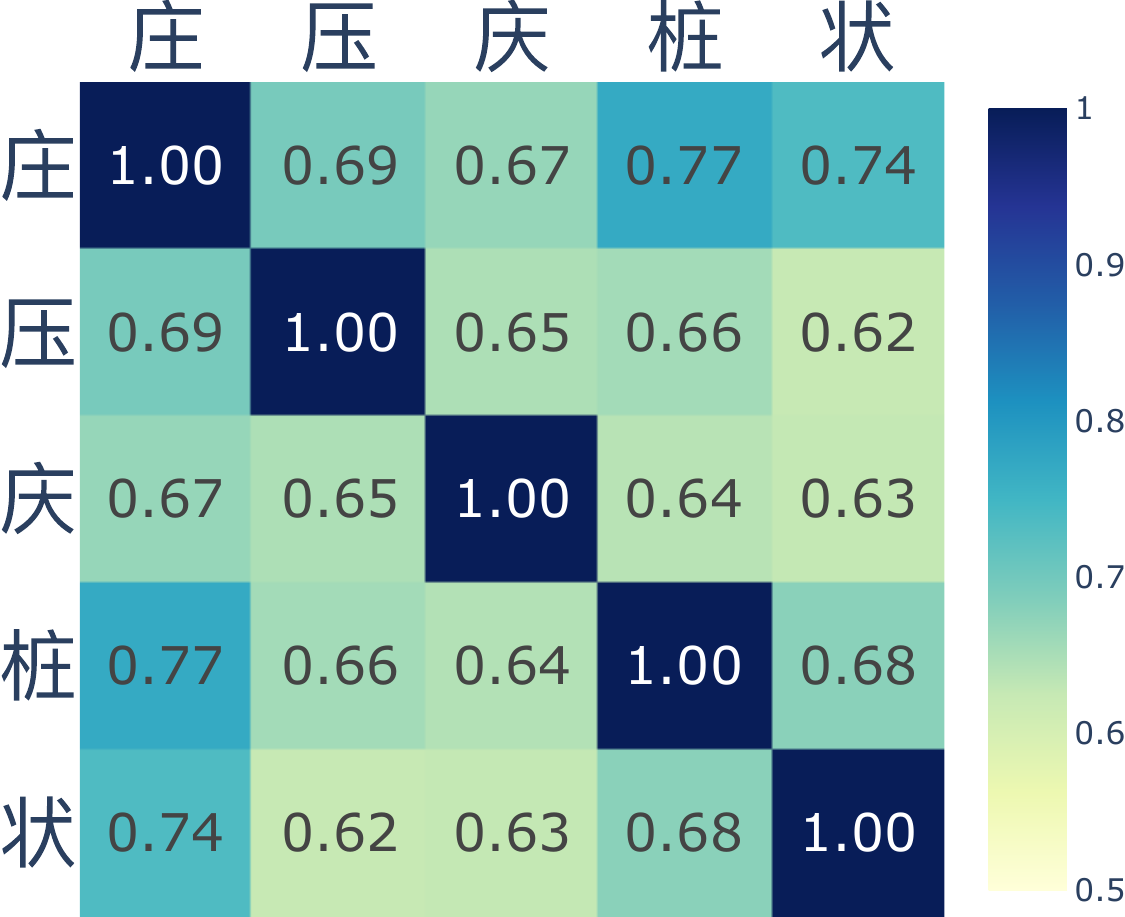}
                \caption{SDCL}
                \label{fig:gull2}
        \end{subfigure}%
        
    \begin{CJK}{UTF8}{gbsn}
  \caption{The cosine similarity between the hidden states of each token in different models. For the sentence ``农民伯伯们正在X稼地里干农活。'', the candidate's characters for ``X'' are ``[庄, 压, 庆, 桩, 状]''. For each model, replace ``X'' with each candidate and get its hidden states, then compute the cosine similarity between these hidden states. And ``庄'' is the correct character. %``压'' and ``庆'' are visually similar to ``庄''; ``桩'', ``状'' are phonetically similar to ``庄''.
  }
  \label{fig:ablate}
  \end{CJK}
        
\end{figure}

\begin{CJK}{UTF8}{gbsn}
  The first one is a case study to show that our model pulls together the last hidden states of different wrong character usages. The results are displayed in Figure \ref{fig:ablate}. As shown, without training, BERT fails to build a connection between the correct character ``庄'' and other characters, and the large cosine similarity between different characters aligns well with \cite{DBLP:conf/emnlp/Ethayarajh19,DBLP:conf/emnlp/GaoYC21}'s observation that the pre-trained word embeddings suffers from anisotropy. 
  The comparison between BERT and SDCL shows that contrastive loss can help BERT better capture the phonological and visual similarity between intra-class characters. 
  %Since for each candidate, SDCL assigns the largest cosine similarity between each candidate and the correct character ``庄''.
\end{CJK}

The second one is the \emph{alignment and uniformity} which is used to measure the quality of representations~\cite{wang2020understanding}. 
With the gold characters as $p_{pos}$ , alignment calculates the expected distance between embeddings of paired characters in the same context.

\begin{equation}
\ell_{\text {align }} \triangleq \underset{\left(x, x^{+}\right) \sim p_{\text {pos }}}{\mathbb{E}}\left\|f(x)-f\left(x^{+}\right)\right\|^2
\end{equation}

On the other hand, uniformity measures how well the embeddings are uniformly distributed:

\begin{equation}
\ell_{\text {uniform }} \triangleq \log \quad \underset{x, y \stackrel{i . i . d}{\sim} p_{\text {data }}}{\mathbb{E}} e^{-2\|f(x)-f(y)\|^2}
\end{equation}

where $p_{data}$ denotes the samples from confusion set\footnote{We use the confusion set realised by \cite{DBLP:conf/acl-sighan/WuLL13}.}.
Specifically, for each sample in the test sets, we replace the source token in the wrong position with a randomly selected token from the confusion set as a negative sample.
As depicted in Table \ref{tb:ablate}, SDCL uniformly learn the embedding and the uniformity loss is reduced at the expense of the elevated alignment loss .

\begin{table}[!ht]
  \setlength{\tabcolsep}{3.3pt}
  \begin{tabular}{lccc}
  \toprule
           & BERT  & SDCL(w/o CL) & SDCL \\
  \midrule
  alignment & 2.58 & 2.62  & \textbf{2.9}      \\
  uniformity & -6.64 & -6.62  & \textbf{-6.88}      \\
  \bottomrule
  \end{tabular}
  \caption{The alignment and uniformity of the model's predictions in the test set.}
  \label{tb:ablate}
\end{table}

% （1）针对测试集中，属于confusion set的词语的性能改善情况。

% （1）测试一下，对于测试集合中句子，每个位置依次替换为对应的confusion chars, 然后计算一下hidden state的cosine similarity 

% 

% \section{Related Work}
% % CSC
% % 对比学习
% % 自蒸馏
% Chinese Spell Checking task is to correct Chinese text that has misspellings. The development of this task has gone through rules-based\cite{chang2015introduction}, machine learning methods\cite{wang2015word}, deep learning methods. Recently, due to advancement of pre-train models, more and more models are built based on them. However, as we have discussed before, BERT is unware of Chinese character's pronounciation and shape. \citet{hong2019faspell} proposed to use the confusion-set to help BERT generate corresponding characters. \citet{ji2021spellbert,zhang2021correcting,li2021exploration,liu2021plome,DBLP:conf/acl/XuLZLWCHM21} tried to narrow the gap through adding phonetic or graphic information during the pre-training phase. Different from all, we utilize the contrastive learning to pull together the hidden states of wrong and right character usages. 

% Contrastive learning tries to pull together postive samples and push away negative samples. 

\section{Conclusion}
%The CSC task needs to determine the correct character usage based on the phonetic and graphic information of the input, while the pre-trained BERT only excels at predicting characters semantically. To fill this gap, we propose using the contrastive loss to pull together these hidden states of the wrong and correct usage of characters, and push away hidden states within the same sentence to avoid anisotropy. Experiments on three CSC datasets reveal that this CL loss is simple and effective. It provides a new perspective on modeling phonological and visual similarity in CSC task.
In this paper, we propose Self-Distillation Contrastive Learning
(SDCL) for CSC task. The proposed method uniform the contextual embedding distribution by contrastive learning with self-distillation. Experiments on three CSC datasets reveal that our method is simple and effective. It provides a new perspective to explore new state-of-the-art results in CSC task.

\bibliography{acl_latex}

\appendix

%\section{Example Appendix}
%\label{sec:appendix}

\section{Data and Metrics}
Following previous work \cite{DBLP:conf/acl/ChengXCJWWCQ20,DBLP:conf/acl/XuLZLWCHM21}, for the evaluation of SIGHAN14 and SIGHAN15, we merge the training set of SIGHAN13\footnote{We add it into the training set just to make sure we use the same training data as previous work.}, SIGHAN14, SIGHAN15 and the generated pseudo data from \citet{DBLP:conf/emnlp/WangSLHZ18}. To make sure our results are comparable with previous work, we directly use the realised processed data from \citet{DBLP:conf/acl/XuLZLWCHM21}\footnote{\url{https://github.com/DaDaMrX/ReaLiSe}}, more details on data processing can be found in \cite{DBLP:conf/acl/XuLZLWCHM21}. For the OCR dataset, we only train on its training set and evaluate in its testing set, and this setting is the same as \cite{hong2019faspell,DBLP:conf/emnlp/JiYQ21}. 

Since the ultimate target of the CSC task is to correct all wrong usages in the sentence, we report the F1, precision and recall metrics in the sentence-level, namely, only when all characters in a sentence are correctly detected\footnote{If the output prediction is not the same token as the input token, we regard our model detect this token as the token need a correction.} or corrected can deem it succeed once.

\section{Implementation Details}
Following \cite{DBLP:conf/acl/XuLZLWCHM21}, we use the pretrained weight realised by \cite{cui2020revisiting}. For all of our models, we use the AdamW optimizer \cite{DBLP:conf/iclr/LoshchilovH19} to optimize our model for 20 epochs, the learning rate is set to be 7e-5, and batch size is 48, $\lambda$ is set to be 0.9 and $\alpha$ is set to be 1 , $\beta$ is set to be 0.05 and $\tau$ is set to be 0.9.

% This is an appendix.

\end{document}